\documentclass{article}

\usepackage{arxiv}
\usepackage[utf8]{inputenc} 
\usepackage[T1]{fontenc}    
\usepackage{hyperref}       
\usepackage{url}            
\usepackage{booktabs}       
\usepackage{amsfonts}       
\usepackage{nicefrac}       
\usepackage{microtype}      
\usepackage{cleveref}       
\usepackage{lipsum}         
\usepackage{float} 
\usepackage{graphicx}
\usepackage{natbib}
\usepackage{doi}

\title{Automatic Prompt Optimization with Prompt Distillation}

\author{\href{https://orcid.org/0009-0000-3865-4446}{\includegraphics[scale=0.06]{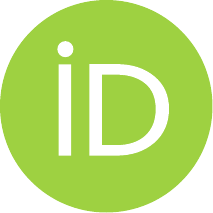}\hspace{1mm}Ernest A. Dyagin} \href{https://orcid.org/0000-0002-3952-6080}{\includegraphics[scale=0.06]{orcid.pdf}\hspace{1mm}Nikita I. Kulin} 
\href{https://orcid.org/0009-0007-0442-7764}{\includegraphics[scale=0.06]{orcid.pdf}\hspace{1mm}Artur R. Khairullin} \href{https://orcid.org/0009-0009-0788-9790}{\includegraphics[scale=0.06]{orcid.pdf}\hspace{1mm}Viktor N. Zhuravlev} \href{https://orcid.org/0009-0002-6046-8943}{\includegraphics[scale=0.06]{orcid.pdf}\hspace{1mm}Alena N. Sitkina}  \\ 
	Computer Technologies Laboratory\\
	ITMO University\\
	Saint-Petersburg, Russia \\
	\texttt{334885@niuitmo.ru 242106@niuitmo.ru 368983@niuitmo.ru} \\
}

\hypersetup{
pdftitle={Automatic Prompt Optimization with Prompt Distillation},
pdfsubject={cs.AI, cs.CL, cs.ML},
pdfauthor={Artur R. Khairullin},
pdfkeywords={LLM, AutoPrompting, Prompt Distillation, Prompting, Prompt Engineering},
}

\begin{document}
\makeatletter
\@ifundefined{undertitle}{\newcommand{\undertitle}{}}{}
\@ifundefined{headeright}{\newcommand{\headeright}{}}{}
\@ifundefined{shorttitle}{\newcommand{\shorttitle}{}}{}
\@ifundefined{shortauthor}{\newcommand{\shortauthor}{}}{}
\makeatother
\maketitle
\begin{abstract}
Autoprompting is the process of automatically selecting optimized prompts for language models, which is gaining popularity due to the rapid development of prompt engineering driven by extensive research in the field of large language models (LLMs). This paper presents DistillPrompt\footnote{Code available as a part of CoolPrompt framework library: https://github.com/CTLab-ITMO/CoolPrompt/}---a novel autoprompting method based on large language models that employs a multi-stage integration of task-specific information into prompts using training data. DistillPrompt utilizes distillation, compression, and aggregation operations to explore the prompt space more thoroughly. The method was tested on different datasets for text classification and generation tasks using the t-lite-instruct-0.1 language model. The results demonstrate a significant average improvement (e.g., 20.12\% across the entire dataset compared to Grips) in key metrics over existing methods in the field, establishing DistillPrompt as one of the most effective non-gradient approaches in autoprompting.
\end{abstract}


\section{Introduction}
In recent years, significant progress has been made in the field of text processing and generation using artificial intelligence—particularly large language models (LLMs) \cite{wei2022finetunedlanguagemodelszeroshot, kadavath2022languagemodelsmostlyknow}. Improving model output quality without modifying its weights falls under the domain of prompt engineering. This field employs various prompting techniques, including Few-shot \cite{brown2020languagemodelsfewshotlearners}, Chain-of-Thought (CoT) \cite{wei2023chainofthoughtpromptingelicitsreasoning}, Directional Stimulus \cite{li2023guiding}, among others. Research indicates that, depending on the task, these techniques can either enhance or significantly degrade model performance. For instance, applying Chain-of-Thought in tasks where reasoning may lead to incorrect answers can result in accuracy drops of tens of percentage points \cite{liu2025mindstepbystep}. Similarly, with Few-shot prompting, it was found that for the Deepseek-R1 model, adding examples to the prompt ``consistently degrades its performance'' compared to a zero-shot task description \cite{deepseekai2025deepseekr1incentivizingreasoningcapability}.

To address this issue, autoprompting methods have emerged—algorithms that leverage both the model itself and various heuristics to automatically improve prompt quality. Studies have shown that prompts generated by these methods often outperform those crafted by humans, even when designed by domain experts \cite{zhou2022large}. It is worth noting that the number of prompting techniques continues to grow each year, making manual prompt engineering increasingly complex and time-consuming. Thus, the challenge of autoprompting remains highly relevant.

This paper presents a novel and more effective non-gradient-based autoprompting approach. The core idea of our method is a complex prompt distillation, which includes: generating diverse prompt candidates, injecting task-relevant examples from a subset of the training data into the prompt, aggregating candidates into a final optimized prompt, and iteratively refining candidates from the final prompt. The proposed approach was evaluated on different datasets and demonstrated superior performance compared to existing non-gradient autoprompting methods.

\subsection{Non-gradient autoprompting methods}

In recent years, there has been significant growth in research on large language models (LLMs) and their applications across various domains. Given that prompting is an integral part of working with these models, numerous studies have explored autoprompting methods.

The first such approach was introduced in \cite{shin2020autopromptelicitingknowledgelanguage}, which relied on fine-tuning an LLM to predict trigger tokens via softmax. However, this method had several limitations as computational overhead, when the LLM required retraining and gradient updates to predict tokens and lack of interpretability, where the generated trigger tokens were not human-interpretable, making it difficult to logically justify their effectiveness, even though most LLMs are inherently black-box models.

Subsequently, non-gradient-based autoprompting algorithms emerged, eliminating the need for gradient updates while allowing the extraction of interpretable prompt patterns for manual refinement \cite{prasad2023gripsgradientfreeeditbasedinstruction, pryzant2023automaticpromptoptimizationgradient}. These approaches leverage semantic parsers, specialized prompt templates, LLMs themselves as the ``brain'' of the autoprompting algorithm. However, prior autoprompting methods suffer from several key drawbacks: insufficient prompt manipulation-limited transformations applied to the prompt structure, randomized instruction selection-arbitrary modification of prompt components without systematic optimization, narrow task applicability-restricted effectiveness across diverse NLP tasks. This paper addresses these limitations by proposing a more structured and generalizable non-gradient approach.

\section{DistillPrompt}

\begin{figure}[H]
    \centering
    \includegraphics[width=1\textwidth]{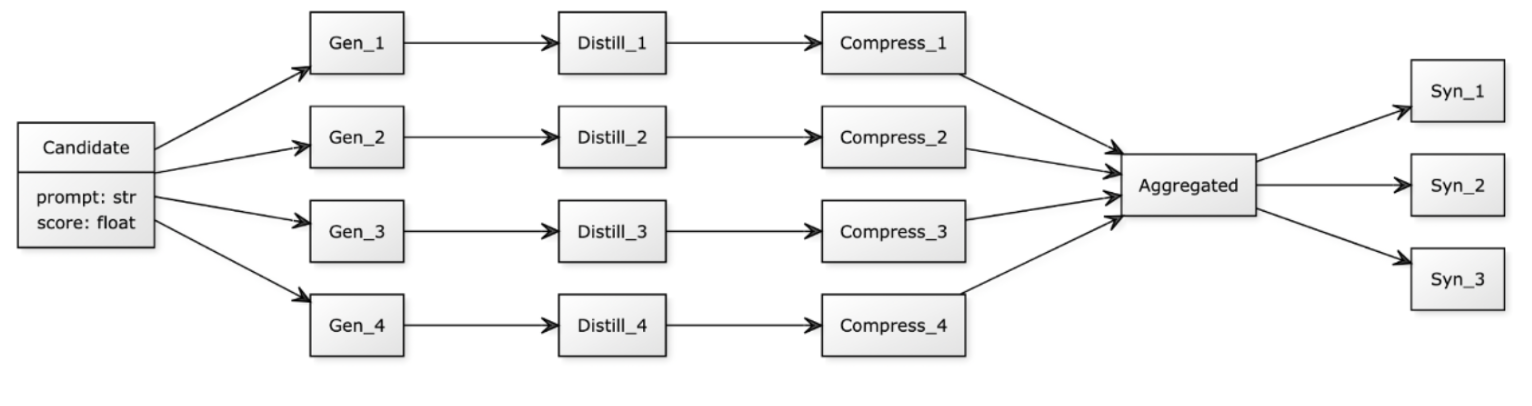}
    \caption{Workflow of DistillPrompt}
    \label{fig:distillworkflow}
\end{figure}

This paper presents an approach that addresses the limitations outlined in the previous section---DistillPrompt, illustrated in Figure \ref{fig:distillworkflow}. The method is based on prompt distillation and incorporates ideas from the Tree-of-Thoughts prompting technique \cite{li2023prompt, yao2023tree}. Prompt distillation refers to manipulating instructions through text compression, reformulating task descriptions, and incorporating usage examples. DistillPrompt is an iterative approach where each iteration consists of five sequential stages.

The initial best candidate is the provided prompt, and in each subsequent epoch, the best candidate from the previous iteration is used, where the best candidate is defined as the one with the highest target metric score on the training set.

At the start of an epoch, variations of the initial prompt are generated. This stage involves creating diverse modifications of the best candidate to explore the task from different perspectives. The purpose is to ``explore'' the space of potentially effective prompts and avoid local optima. As a result, N new prompt candidates are produced (N=4 in the current implementation). The number of candidates is a hyperparameter of the algorithm, balancing the trade-off between the number of large language model (LLM) calls and the coverage of the prompt space for the given task. Generation is performed via queries to the LLM with a temperature of 0.7 to enhance creativity while minimizing the risk of generating uninterpretable prompts.

The next stage is example embedding. While the previous step yielded four new prompt candidates, they explore the prompt space ``blindly''. To guide them toward the target task while preserving their unique formulations, we propose embedding examples from the training set. Initially, we tested direct example insertion (as in one-shot and few-shot techniques), but this proved less effective than using the LLM to analyze examples and extract their underlying task-solving principles, which better captures task-relevant information. For each prompt candidate, K examples are independently and randomly selected from the training set (K=5 in this implementation) to guide the LLM in refining the prompt. However, there is a risk of the LLM ``overfitting'' to the examples-focusing on their specific labels and questions rather than deriving generalizable insights.

To mitigate this, the next stage involves instruction compression, where the LLM condenses the prompts from the previous step into a few sentences retaining the core ideas introduced by the examples and the overarching task objective. This step helps generalize the prompts while preserving the insights gleaned from the examples.

Next, candidate aggregation is performed. Since examples were selected independently and randomly for each candidate, the extracted insights vary. Thus, the natural progression is to merge the compressed candidates into a single distilled prompt encompassing the collective ideas.

The final stage generates new candidates from the distilled prompt by creating variations (as in Stage 1). The resulting candidates are evaluated on tasks, and the top-performing candidate becomes the new initial prompt for the next epoch until the epoch limit is reached. Since the process explores the prompt space, candidates may outperform or underperform those from previous epochs; thus, the method requires multiple iterations. The algorithm's output is the best prompt from the final epoch.

\section{Experimental Evaluation}
\label{sec:evaluation}

\subsection{Experimental Setup}
\label{subsec:exp_setup}

To validate the effectiveness of DistillPrompt, we designed the following experimental setup: each evaluated method was tested across multiple datasets using the t-lite-instruct-0.1 LLM. This paper introduces a comprehensive benchmark (a curated dataset collection) for comparing non-gradient autoprompting methods, including the proposed solution.

For a robust evaluation framework, we analyzed multiple autoprompting studies \cite{pryzant2023automaticpromptoptimizationgradient, li2023prompt, yao2023tree}, to compile diverse datasets. The resulting benchmark encompasses both classification and question-answering tasks, along with various text generation challenges. The complete dataset list is: SST-2, MedQA, GSM8K, MNLI, MR, TREC, SAMSum, BBH (BIG-Bench Hard). In this benchmark, question-answering datasets involve multiple-choice responses (similar to exam questions), making them effectively multiclass classification tasks. The benchmark datasets can be broadly categorized into classification and generation tasks, each requiring specific evaluation metrics.

While many autoprompting studies use accuracy for classification evaluation, its simplicity fails to capture class distribution nuances. Therefore, we employ macro F1-score as our primary classification metric. For generation tasks, we selected METEOR \cite{banerjee2005meteor} - an F1-analog metric that measures word-level precision and recall (with higher recall weighting), making it particularly suitable for text generation evaluation. The baseline comparisons include prompting techniques (baseline prompt---original dataset-provided prompt, few-shot prompt---baseline prompt augmented with three training examples) and non-gradient autoprompting approaches (Grips, Protegi). This experimental design enables systematic comparison of DistillPrompt against both manual prompting techniques and state-of-the-art autoprompting methods across diverse NLP tasks.

\subsection{Results}

The experimental results across metrics and datasets are presented in Tables \ref{tab:cls_results} and \ref{tab:gen_results} for the t-lite-instruct-0.1 model, showing performance on classification and generation tasks respectively. The BBH metric values in the tables represent averages across all tasks from the original benchmark. Notably, Protegi was excluded from Table \ref{tab:gen_results} as its methodology is not adapted for generation tasks.

\begin{table}[H]
\centering
\caption{Results of DistillPrompt on classification tasks with t-lite-instruct-0.1}
\label{tab:cls_results}
\begin{tabular}{lcccccc}
\toprule
\textbf{Method} & \textbf{sst-2, f1} & \textbf{mnli, f1} & \textbf{trec, f1} & \textbf{mr, f1} & \textbf{medqa, f1} & \textbf {bbh, f1} \\
\midrule
Baseline prompt & 0.6135 & 0.4178 & 0.28673 & 0.8617 & 0.2957 & 0.2055 \\
Few shot: n = 3 & 0.9328 & 0.3741 & 0.2681 & 0.6031 & 0.2397 & 0.3129 \\
Protegi & 0.6397 & 0.4964 & \textbf{0.3555} & 0.6363 & 0.2935 & 0.3718 \\
Grips & 0.6135 & 0.7407 & 0.3153 & 0.9117 & \textbf{0.3032} & 0.2879 \\
\textbf{DistillPrompt: v1.0 (ours)} & \textbf{0.9484} & \textbf{0.7606} & 0.3526 & \textbf{0.9392} & 0.2957 & \textbf{0.4045} \\
\bottomrule
\end{tabular}
\end{table}

\begin{table}[H]
\centering
\caption{Results of DistillPrompt on generation tasks with t-lite-instruct-0.1}
\label{tab:gen_results}
\begin{tabular}{lcccc}
\toprule
\textbf{Method} & \textbf{gsm8k, METEOR} & \textbf{samsum, METEOR} & \textbf{bbh, METEOR} \\
\midrule
Baseline prompt & 0.02932 & 0.44787 & 0.1247 \\
Few shot: n = 3 & 0.0179 & 0.38484 & 0.2100 \\
Grips & 0.02643 & 0.45516 & 0.1491 \\
\textbf{DistillPrompt: v1.0 (ours)} & \textbf{0.0347} & \textbf{0.4579} & \textbf{0.2961} \\
\bottomrule
\end{tabular}
\end{table}

\section{Discussion}
\label{sec:discussion}

The conducted experiments demonstrate that DistillPrompt effectively handles both classification and text generation tasks. Across all evaluated datasets, DistillPrompt either outperformed or matched the performance of existing non-gradient autoprompting methods with a significant average improvement 20.12\% across the entire dataset compared to Grips. 

For classification tasks, the average F1-score improved by 36.18\% comparing baseline prompt and by 15.09\% comparing the strongest of baselines---Grips. In text generation tasks, the average METEOR score increased by 31.03\% comparing baseline prompt and by 25.05\% comparing the strongest of baselines - Grips. Comparisons and improvements were calculated relative to the maximum average metrics achieved by existing solutions.

This work creates a scope for future research into distillation of prompts and other non-gradient autoprompting methods. The current DistillPrompt implementation could potentially be further refined for more targeted prompt optimization. Moreover, the concept of prompt distillation could be generalized and adapted to other non-gradient autoprompting methods, representing a promising direction for future studies.

\section{Conclusion}
\label{sec:conclusion}

The proposed DistillPrompt algorithm, which employs distillation prompt technique for prompt optimization, was evaluated on classification and generation datasets covering various natural language processing domains. It demonstrated consistent improvements over existing non-gradient algorithm-based autoprompting methods. DistillPrompt proves to be a competitive solution, showing that exploring prompt distillation for autoprompting can yield significant benefits and advance current methods to new levels of performance.

\bibliographystyle{plain}
\bibliography{references}

\end{document}